# A Comparative Assessment of Concrete Compressive Strength Prediction at Industry Scale Using Embedding-Based Neural Networks, Transformers, and Traditional Machine Learning Approaches

Md Asiful Islam[1], Md Ahmed Al Muzaddid[2], Afia Jahin Prema[3], Sreenath Reddy Vuske[4]

**Corresponding Author:** Md Asiful Islam (Email: asiful.islam.me@gmail.com)

[1] Intertek PSI, Dallas, TX 75243, USA (e-mail: asiful.islam.me@gmail.com, mdasiful.islam@intertek.com).

[2] University of Texas at Arlington, Arlington, TX 76010, USA (e-mail: mdahmedal.muzaddid@mavs.uta.edu).

[3] University of Texas at Arlington, Arlington, TX 76010, USA (email: jahinprema2014@gmail.com)

[4] Intertek PSI, Dallas, TX 75243, USA (e-mail: sreenath.vuske@intertek.com).

## Abstract

Concrete is the most widely used construction material worldwide; however, reliable prediction of compressive strength remains challenging due to material heterogeneity, variable mix proportions, and sensitivity to field and environmental conditions. Recent advances in artificial intelligence enable data-driven modeling frameworks capable of supporting automated decision-making in construction quality control. This study leverages an industry-scale dataset consisting of approximately 70,000 compressive strength test records to evaluate and compare multiple predictive approaches, including linear regression, decision trees, random forests, transformer-based neural networks, and embedding-based neural networks. The models incorporate key mixture design and placement variables such as water–cement ratio, cementitious material content, slump, air content, temperature, and placement conditions. Results indicate that the embedding-based neural network consistently outperforms traditional machine learning and transformer-based models, achieving a mean 28-day prediction error of approximately 2.5%. This level of accuracy is comparable to routine laboratory testing variability, demonstrating the potential of embedding-based learning frameworks to enable automated, data-driven quality control and decision support in large-scale construction operations.

## 1. Introduction

Concrete compressive strength (CCS) plays a critical role in structural design, construction scheduling, and long-term performance evaluation of the structure. As a composite material, concrete's mechanical behavior is influenced by a wide range of factors: cement type, water-cement ratio, aggregate characteristics, admixtures, and curing conditions [1]. Predicting compressive strength from these parameters remains a complex challenge due to nonlinear interactions and material variability, particularly in field applications. Traditional empirical models and statistical regression techniques have long been used to estimate compressive strength [2],[3]. While useful in controlled scenarios, these methods often fail to generalize across varying mix designs and environmental conditions. Early studies exploring artificial neural networks (ANNs) demonstrated promise in capturing nonlinear dependencies. Ni et al. [4], for example, applied a basic feed-forward neural network to predict CCS and highlighted its advantages over linear models. Subsequent studies explored more advanced computational techniques, such as fuzzy logic and hybrid systems [5]-[8], but these approaches were generally validated on relatively small datasets of only a few hundred samples, limiting their broader applicability. Deepa et al.[8] compared multiple algorithms including linear regression, multi-layer perceptron (MLP), and M5P trees, based on data collected from a real-world construction setting. While their study offered valuable insights into model performance, the dataset comprised only 300 samples, limiting statistical robustness. Similarly, Chopra et al. [9],[10] evaluated regression and tree-based models on relatively small data pools. Chou et al. [11] conducted a critical review of machine learning applications in concrete strength prediction; however, the majority of existing models were developed using narrowly scoped experimental datasets. Some studies have attempted classification-based approaches rather than regression-based approaches to predict CCS, such as categorizing concrete strength into low, medium, and high bins [12]. While such approaches may suit quality screening, they fall short of providing precise compressive strength values necessary for performance-based specifications. Asteris and Mokos [13] developed a neural network using only two input features derived from non-destructive tests. Though innovative, the model's input limitation and narrow use case limit its broader applicability. In recent years, a range of machine learning techniques, including gradient boosting, random forest, support vector machines, and hybrid ensemble methods, have been applied to concrete compressive strength prediction [14]–[20]. Machine learning models have been applied to concrete compressive strength prediction across a variety of material systems, including high-performance concrete [21]–[26] and lightweight self-compacting concrete [27]. These studies demonstrate the potential of data-driven approaches to capture nonlinear relationships between mixture design parameters and strength development. At the same time, existing methods exhibit different tradeoffs related to model complexity, computational demand, and interpretability. For example, support vector machine models typically require careful kernel selection and parameter tuning, decision tree-based models can be sensitive to data partitioning, and ensemble methods often prioritize predictive accuracy over transparent model logic. These characteristics highlight the need for systematic, large-scale comparisons of predictive approaches under consistent evaluation conditions. The predictive approaches proposed in this paper aim to address several of these limitations. First, it uses a large-scale dataset of over 70,000 concrete cylinder break records obtained from a major U.S. infrastructure project, ensuring wide variability in mix types, placement conditions. This dataset reflects real world field conditions and material diversity far beyond typical laboratory settings. Second, the study evaluates five predictive models side-by-side: Decision Tree, Random Forest, Multiple Linear Regression, Artificial Neural Network with Embedding, and a Transformer-based architecture. This comparison offers insights into the relative strengths of classical, ensemble, and deep learning methods on field-scale data. Notably, this study represents one of the first applications of embedding-based neural networks that integrate both categorical and numerical variables for predicting concrete compressive strength (CCS). In addition, we employed a Transformer model to predict CCS using large-scale field data. Originally introduced by Vaswani et al. [28], the Transformer architecture revolutionized natural language processing through its self-attention mechanism and highly parallelizable computation. Unlike sequential models such as LSTMs [29], Transformer-based models provide a flexible framework for analyzing tabular datasets that include both categorical and continuous variables. When combined with embedding layers for categorical inputs, such as material codes and fracture type, these models are able to represent complex feature interactions within structured construction data. In this study, model performance is evaluated using the coefficient of determination ($R^2$), mean absolute error (MAE), and mean absolute percentage error (MAPE), together with feature importance analysis across models. The comparative results highlight differences in predictive capability among modeling approaches and support the identification of architectures that are most suitable for large-scale concrete quality assurance applications. The findings also inform future research directions related to data-driven construction analytics and potential integration with digital quality control frameworks.

## 2. Experimental Framework and Sample Preparation

A comprehensive dataset comprising approximately 70,000 concrete compressive strength test records was collected over a four-year period (May 2021–May 2025) from a large-scale design-build infrastructure project in the United States. The data originated from routine quality assurance testing conducted in accordance with Project requirements and ASTM standards by the certified field and lab technicians, under the oversight of quality assurance engineers. These tests formed part of the standard construction material evaluation process and were tightly integrated with site operations, capturing a wide variety of concrete mix designs and field conditions. Concrete samples were cast using 142 distinct mix designs, used across 48 different structural feature applications starting from bridge superstructures, concrete paving and non-structural operations like curb and riprap placements. The cylinder preparation and testing workflow is shown in Fig.1.

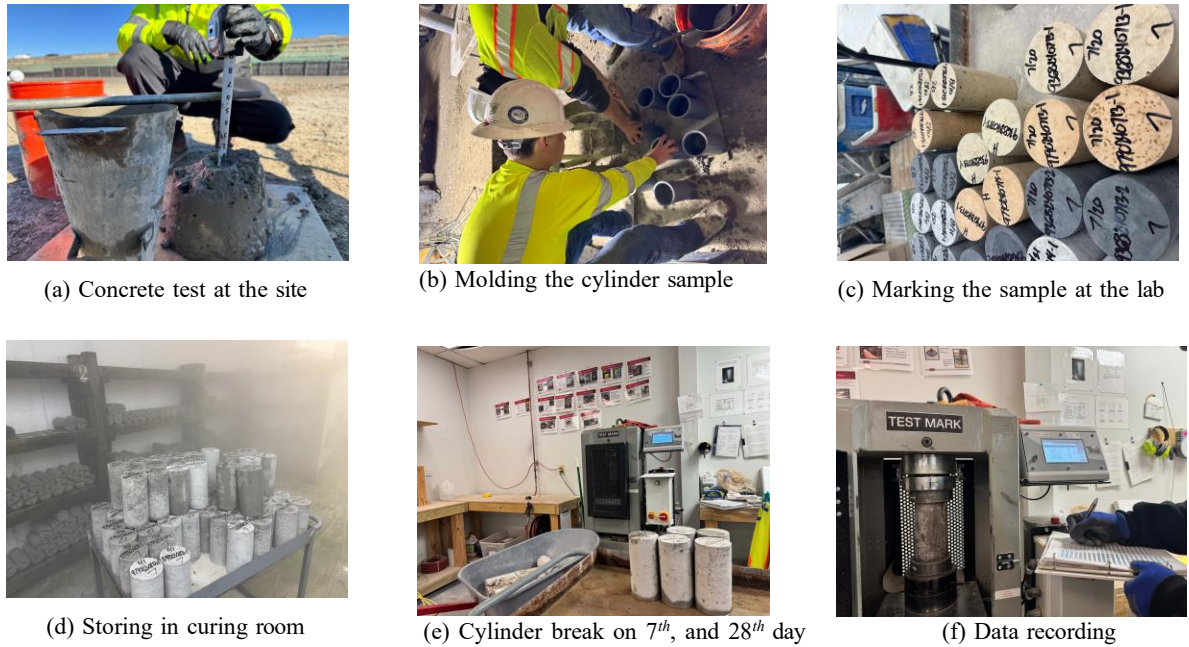

(a) Concrete test at the site  (b) Molding the cylinder sample  (c) Marking the sample at the lab

(d) Storing in curing room  (e) Cylinder break on $7^{th}$, and $28^{th}$ day  (f) Data recording

Fig. 1. Concrete cylinder testing workflow

The concrete was batched primarily at three plants dedicated to the project. The samples were tested at 7-day and 28-day intervals, with 56-day tests conducted selectively when verification or dispute resolution was required. Testing and sample preparation adhered strictly to ASTM C31 procedures. Each 4 in. × 8 in. cylinder was prepared in two lifts, with each lift subjected to 25 rodding actions and 12–15 tamps to ensure proper consolidation. The molds were made by the certified ACI Concrete field testing technicians. Calibrated slump cones, thermometers, and air meters were used consistently during sampling to capture fresh plastic concrete properties. Water cement ratio and other batching parameters were verified at the site before making the sample molds. Following molding, the cylinders were stored in insulated containers (maintained at 60°F to 80°F) for 12–24 hours at the field site before being transferred to the central project laboratory. Upon arrival, samples were demolded, marked for identification, and placed in a controlled curing room with 100% relative humidity and a temperature of 70°F to 77°F. This ensured compliance with standard curing protocols and minimized moisture loss. Strength testing was performed using a TestMark 400K automatic break machine, operated by ACI-certified strength technicians. The break machine underwent annual calibration. Two cylinders from each sample set were tested at the designated age, and the results were averaged to ensure consistency. When discrepancies or high variability were observed, additional hold cylinders were tested at 56 days, and a secondary accredited laboratory was engaged to test approximately 20% of the total samples, providing an assessment of inter-laboratory variability and reinforcing confidence in the reported data. Through rigorous documentation, strict calibration protocols, and systematic verification, measurement error and procedural variability were minimized, resulting in high-quality data well- suited for machine learning model development. Each mix design underwent a rigorous review and approval process by a licensed professional engineer, ensuring compliance with project-specific requirements as well as the governing specifications and guidelines established by the client and relevant regulatory authorities. The primary constituents of these concrete mixtures comprised portland cement, supplementary cementitious materials such as fly ash, coarse and fine aggregates, potable water, and a range of chemical admixtures. Among the admixtures used were water-reducing agents, high-range water reducers (superplasticizers), air-entraining agents, and set retarders, all incorporated to tailor the fresh and hardened properties of the concrete to the performance expectations of the project. Table 1 presents the acceptable gradation ranges for coarse aggregate as determined through sieve analysis.

**Table 1**: Gradation Limits for Coarse Aggregate Based on Sieve Analysis.

| Sieve Size | Cumulative Passing (%) |
|---|---|
| 2-1/2" | 100 |
| 2" | 100 |
| 1-1/2" | 95–100 |
| 1" | 95–100 |
| 3/4" | 60–90 |
| 1/2" | 25–60 |
| #4 | 0–10 |
| #8 | 0–5 |

These limits, established in accordance with project specifications and ASTM C33 requirements, ensure that aggregate distribution supports proper workability, minimizes segregation, and enhances inter-locking within the concrete matrix. Maintaining the specified ranges across all sieve sizes was critical for consistent strength development and durability throughout the project duration. Table 2 illustrates the gradation requirements for fine aggregates (concrete sand).

**Table 2**: Gradation Limits for fine aggregate-based on sieve analysis.

| Sieve Size | Cumulative Passing (%) |
|---|---|
| 3/8" | 100 |
| #4 | 95–100 |
| #8 | 80–100 |
| #16 | 50–85 |
| #30 | 25–65 |
| #50 | 10–35 |
| #100 | 0–10 |
| #200 | 0–3 |

Adherence to these limits controlled the fineness modulus and ensured adequate packing density, thereby improving paste–aggregate bonding, reducing voids, and enhancing workability. Regular monitoring of sand gradation confirmed compliance with project standards and contributed to achieving the targeted performance in fresh and hardened concrete. Strict adherence to these requirements ensured uniformity in aggregate quality, which is widely recognized as a crucial factor in controlling the mechanical properties, workability, and long- term durability of concrete. The following parameters in Table 3 were selected as input variables for this study, as they are widely recognized to influence the compressive strength and overall performance of concrete. These include material proportions, fresh concrete properties, and environmental factors at the time of placement. Table 3 presents the statistical summary of each parameter, including unit, minimum, maximum, mean, approximate count, and standard deviation, derived from approximately 70,000 samples. This dataset captures the variability in cementitious content, water–cement ratio, placement conditions, and mixture characteristics, thereby providing a robust foundation for evaluating the relative significance of these factors on strength development.

**Table 3**: Statistical summary of input parameters used in concrete strength analysis.

| Parameter | Unit | Min | Max | Mean | Count (Approx.) | Standard Deviation |
|---|---|---|---|---|---|---|
| Unit Cement Material | lb/CY | 208.5 | 1380.0 | 571.95 | 70,000 | 73.67 |
| Water-Cement Ratio (WC) | N/A | 0.14 | 3.03 | 0.38 | 70,000 | 0.06 |
| Concrete Temperature | F | 50.0 | 97.0 | 76.0 | 70,000 | 8.96 |
| Air Temperature | F | 35.0 | 109.0 | 71.1 | 70,000 | 14.97 |
| Elapsed Time | Min | 5.0 | 118.0 | 39.7 | 70,000 | 13.43 |
| Placement Air | % | 1.0 | 9.5 | 4.3 | 70,000 | 1.1 |
| Placement Slump | Inch | 1.0 | 9.0 | 5.5 | 70,000 | 1.5 |
| Material Code | Unique | | | | 142 | |
| Fracture Type | Unique | | | | 6 | |

### 3. Input Parameters:

This chapter describes the key variables used to develop and evaluate the predictive models. The selected parameters reflect material composition, fresh concrete properties, placement conditions, and observed fracture characteristics that are routinely captured during field quality assurance testing and are known to influence compressive strength development. Each variable is briefly described in the following subsections to provide physical context, engineering relevance, and justification for its inclusion in the modeling framework. Together, these parameters form the basis for the comparative evaluation of traditional machine learning and deep learning approaches presented in the subsequent sections.

*3.1 Unit Cementitious Material (lb/CY)*

This parameter denotes the cementitious material content batched per cubic yard of concrete, expressed in pounds per cubic yard of concrete. It includes both cement and supplementary cementitious materials such as fly ash. The cementitious content is a critical variable influencing not only the compressive strength but also the durability and economic aspects of concrete. While increased cementitious content generally contributes to improved early-age and long-term strength, it can simultaneously induce higher heat of hydration, elevate the risk of drying shrinkage, and raise production costs. In the present investigation, the majority of concrete placements utilized Type I-L portland-limestone cement, with a smaller proportion incorporating traditional Type I portland cement. Fly ash was introduced as an SCM in selected mixtures, with both Class F and Class C fly ash employed during different periods of batching, reflecting variations in material availability and project requirements [30].

### 3.2 Water-Cement Ratio (W/C)

The water-cement ratio (W/C), defined as the quotient of the total mass of water divided by the mass of cementitious materials, is widely recognized as one of the most influential parameters governing the performance of concrete. A lower W/C ratio generally enhances compressive strength and reduces permeability, thereby contributing to improved durability and long-term serviceability [31]. Conversely, excessively low ratios may compromise workability, often necessitating the incorporation of chemical admixtures to ensure adequate placement and consolidation. In practice, mixing water may originate from three primary sources: (i) free moisture associated with coarse and fine aggregates, (ii) batched water measured and introduced at the plant, and (iii) supplementary water occasionally added at the construction site prior to discharge to achieve the desired plastic properties of the mix.

### 3.3 Concrete Temperature (°F)

The temperature of freshly mixed concrete at the time of placement exerts a significant influence on hydration kinetics, workability, and the rate of setting. Elevated concrete temperatures accelerate the hydration process, leading to rapid strength development at early ages; however, they simultaneously heighten the risk of thermal cracking, increased evaporation-induced plastic shrinkage, and potential long-term durability concerns. In contrast, lower placement temperatures reduce the rate of hydration, which may extend setting time and delay early-age strength gain, potentially impacting construction schedules.

### 3.4 Air Temperature (°F)

Ambient air temperature during placement plays a significant role in the curing and performance of fresh concrete. Extreme variations in air temperature can impact hydration rates and strength development. Without proper curing practices, hot conditions can lead to rapid moisture loss, while cold conditions may delay hydration and risk early freezing of the concrete, adversely affecting strength development.

### 3.5 Elapsed Time (minutes)

Elapsed time refers to the period between batching and placement of concrete. Managing this interval is critical to ensuring workability and maintaining the integrity of the prepared sample. Prolonged elapsed times may reduce slump, hinder consolidation, and in some cases initiate premature hydration. Admixtures such as water reducers or retarders are often employed to mitigate these effects when extended transport or placement times are anticipated.

### 3.6 Placement Air Content (%)

The air content of fresh concrete represents the percentage of entrained or entrapped air. Properly controlled air entrainment is essential for improving freeze–thaw resistance and mitigating scaling in environments subject to temperature cycling. However, excessive air content can reduce compressive strength, as voids within the hardened matrix reduce load-bearing capacity. Optimal ranges, typically governed by ASTM C231 or ASTM C173, were maintained to balance durability and strength.

### 3.7 Placement Slump (inches)

Slump, measured in accordance with ASTM C143, provides an indication of the concrete's consistency and workability. Adequate slump ensures proper placement, consolidation, and surface finishing without segregation or bleeding. Slump values are influenced by W/C ratio, admixture usage, and aggregate gradation. Maintaining slump within the specified range is critical for ensuring both constructability and performance.

### 3.8 Material Code

Each concrete batch was assigned a unique material code, enabling traceability and quality control across the project. The code identifies critical mix characteristics, including the plant source, concrete class, percentage of fly ash substitution, cement type, and whether the mix was designated as high-performance concrete. For example, the material code 1-C20HPC-1L corresponds to Plant 1, Class C concrete, a 20% fly ash replacement, High Performance Concrete, and Type 1L cement.

### 3.9 Fracture Type

Fracture type classification was performed based on failure modes observed during compressive strength testing. In line with ACI 214R guidelines, fracture types provide an indication of the validity of test results and can reveal potential material or procedural deficiencies. Six types of failure modes were recorded, namely cone, cone with vertical cracks, columnar vertical cracks, diagonal shear, edge fracture, and edge fractures occurring on one end. Typical patterns, such as cone or shear fractures, are generally considered valid, while irregular or atypical fractures may indicate issues such as improper consolidation, inadequate end preparation, or equipment misalignment. Documenting fracture types enhances the reliability of strength data, supports quality assurance evaluations, and aids in resolving disputes when failures occur.

## 4. Analysis and Modeling

This section presents the analytical framework and predictive models evaluated in this study. The raw dataset was randomly partitioned into training and testing subsets using an 80–20 split, where 80% of the data were used for model training and 20% were reserved for independent performance evaluation. Five modeling approaches were examined, including multiple linear regression, decision tree, random forest, a transformer-based neural network, and an embedding-based neural network. All models were trained using the same input parameters to ensure a consistent and fair comparison. Model performance was assessed using the coefficient of determination ($R^2$), mean absolute error (MAE), and mean absolute percentage error (MAPE). In addition, feature importance analysis was conducted, where applicable, to examine the relative contribution of mixture design, placement, and environmental variables to compressive strength prediction and to support interpretation of model behavior.

### 4.1 Multiple Linear Regression

Multiple Linear Regression (MLR) is a statistical modeling technique used to examine the relationship between a dependent variable (also known as the target variable) and multiple independent variables (predictors). The objective of MLR is to model the target variable as a linear combination of the predictors, allowing us to predict its value based on the input features. The general form of the MLR equation is:

$$y = \beta_0 + \beta_1 x_1 + \beta_2 x_2 + \cdots + \beta_n x_n + \varepsilon \qquad (1)$$

where:
$y$ = Dependent variable (target)
$\beta_0$ = Intercept of the regression line
$\beta_1, \beta_2, ..., \beta_n$ = Coefficients of the independent variables
$x_1, x_2, ... x_n$ = Independent variables (predictors)
$\varepsilon$: Error term (residuals) accounting for variability not explained by the model

*4.2 Decision Tree*

Decision Tree is a supervised learning algorithm that models data by learning simple decision rules inferred from input features [32]. It functions by splitting the dataset into branches based on conditions that maximize the separation between target outputs, typically using metrics such as Gini impurity or information gain. Each internal node represents a decision on a feature, and each leaf node corresponds to a predicted output. Its intuitive tree-like structure makes it highly interpretable and easy to visualize. However, Decision Trees tend to overfit, especially on complex or noisy datasets, unless regularized through techniques such as pruning or limiting tree depth. Despite its simplicity, Decision Tree provides a solid baseline for regression tasks, especially when combined with ensemble techniques like Random Forest.

*4.3 Random Forest*

Random Forest is a robust and versatile machine learning algorithm widely used for classification and regression tasks. It operates by constructing a multitude of decision trees during training and aggregating their results to make a final prediction [33]. Each tree in the forest is built on a random subset of the data and selects a random subset of features for splitting nodes. This randomness reduces overfitting and enhances the model's ability to generalize to unseen data. In regression, the Random Forest outputs the average prediction of individual trees, while in classification, it chooses the class with the majority vote. A graphical structure of the Random Forest model is shown in Fig. 2. This ensemble approach makes Random Forest highly accurate, resistant to overfitting, and capable of handling large datasets and high-dimensional feature spaces.

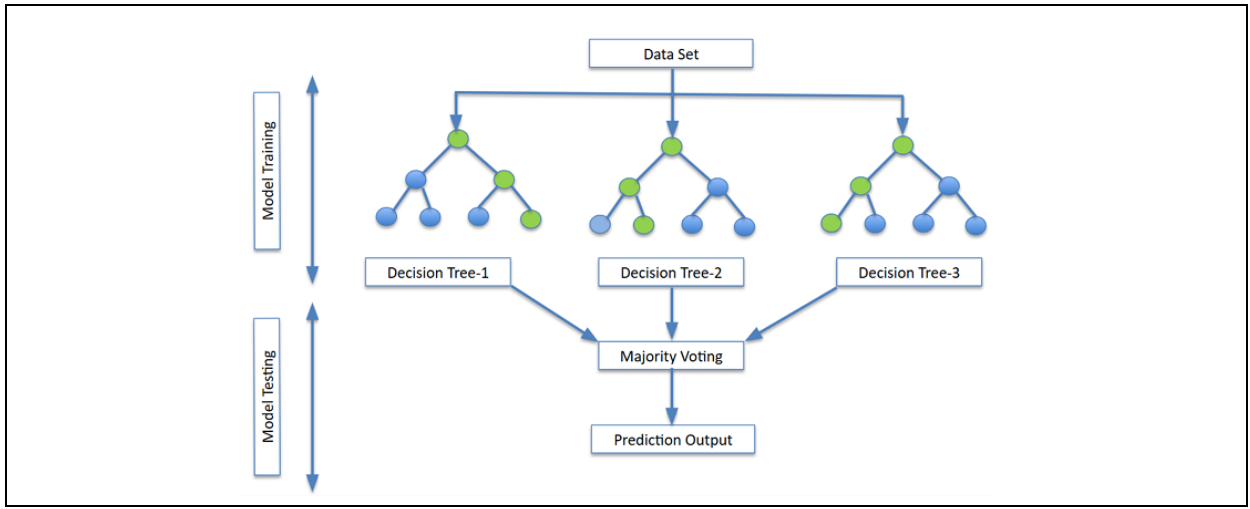

Fig. 2. Random Forest ensemble of decision trees with aggregated prediction.

*4.4 Transformer-based Model*

Transformer architectures have gained attention beyond natural language processing due to their ability to model complex interactions in structured datasets. In this study, a customized Transformer encoder was adapted for regression on tabular concrete data. The Transformer model used in this study is a customized deep learning architecture inspired by the original Transformer encoder design. Unlike models developed for natural language tasks, this implementation reshapes the tabular input features into a format that allows the attention mechanism to evaluate relationships across all feature dimensions. Multi-head self-attention layers enable the model to assign different weights to each input feature based on its contextual relevance, while residual connections and layer normalization help stabilize training. Embedding layers were not applied, as categorical variables were preprocessed using one-hot encoding. The model was trained using a regression objective and evaluated based on the same metrics. Its ability to learn long-range dependencies across input features makes it especially suited for capturing hidden interactions in large, complex datasets like the one used in this study [28].

*4.5 Embedding-based NN*

The embedding-based model is depicted in Fig. 3. The model begins by accepting two types of inputs: numerical and categorical features. The numerical inputs consist of real-valued features e.g. Unit Cement Material, Water-Cement Ratio, Concrete Temperature etc. For categorical inputs, i.e. Material Code and Feature, unique values of each variable are mapped to an arbitrary index and then passed through a dedicated embedding layer. The size of each embedding is determined by $2[\log_2 K]$, where K is the number of unique values for each categorical variable. These embedding layers are initialized with weights drawn from a uniform distribution in the range $[-1, 1]$. Their role is to transform sparse categorical variables into dense vector representations that encode latent relationships among categories. Once the embeddings are obtained, they are concatenated with the numerical inputs to form a unified feature vector. This combined vector undergoes layer normalization using LayerNorm to stabilize the input distribution and improve training convergence. The normalized vector is then passed through a sequence of five fully connected (FC) blocks that form the hidden layers of the model. Each hidden block comprises a linear transformation followed by a layer normalization and a GELU (Gaussian Error Linear Unit) activation function. All linear layers are initialized using Kaiming uniform initialization, which is optimized for ReLU-based activations but is compatible with GELU as used here. The final layer of the model is a regression head, which consists of a linear transformation from the previous hidden dimension to a single output neuron, followed by a ReLU activation. This setup ensures that the predicted concrete strength remains non-negative. In terms of architectural components, the model contains one embedding layer for each categorical feature, six-layer normalization operations (one at the input and five within the hidden blocks), and seven linear layers (five in the hidden layers, one in the output layer, and optionally within embedding projections). It includes six activation layers in total—five GELU and one ReLU, leading to a total of 24 sublayers or modules, not counting the embedding layers.

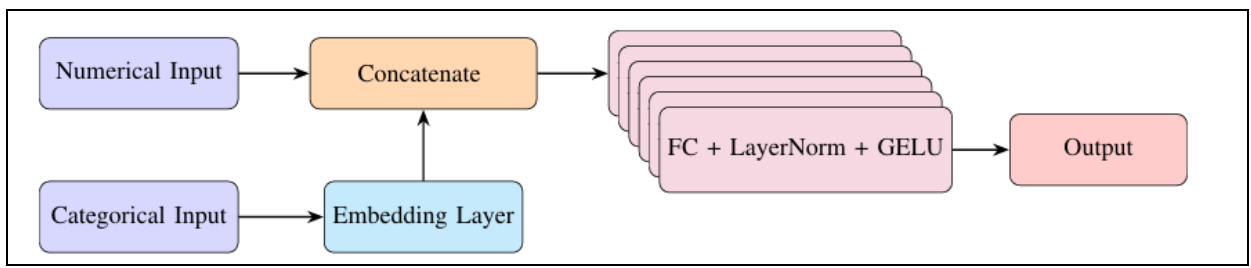

Fig. 3: Embedding-based neural network architecture.

## 5. Results and Discussion

Table 4 presents the comparative performance of the evaluated models in predicting concrete compressive strength at both 7-day and 28-day curing ages. A clear variation in predictive accuracy is observed across the methods, highlighting the influence of model architecture and feature representation on predictive capability. The Feature Embedding-based Neural Network (NN) demonstrated the highest accuracy among all models, achieving R² values of 0.87 and 0.90 at 7-day and 28-day, respectively, with the lowest corresponding MAE (160.90 psi and 143.09 psi) and MAPE (3.37% and 2.50%). This superior performance can be attributed to its ability to capture nonlinear interactions and latent feature relationships that conventional models fail to address. The embedding layer effectively encoded categorical and numerical variables, enabling the network to learn complex dependencies such as the synergy between cementitious materials, admixture dosages, and curing conditions.

**Table 4**: Model performance comparison (7-Day and 28-Day predictions)

| Method | $R^2$ | | MAE (psi) | | MAPE (%) | |
|---|---|---|---|---|---|---|
| | 7-Day | 28-Day | 7-Day | 28-Day | 7-Day | 28-Day |
| Linear Regression | 0.59 | 0.61 | 556.06 | 653.66 | 11.68 | 10.67 |
| Decision Tree | 0.81 | 0.79 | 212.69 | 271.69 | 4.50 | 4.30 |
| Random Forest | 0.85 | 0.86 | 296.57 | 348.95 | 6.28 | 5.64 |
| Transformer-based NN | 0.66 | 0.69 | 500.08 | 574.24 | 10.37 | 9.22 |
| Feature Embedding-based NN | 0.87 | 0.90 | 160.90 | 143.09 | 3.37 | 2.50 |

In comparison, Random Forest also performed well, with $R^2$ values of 0.85 and 0.86. Its ensemble nature allowed it to generalize effectively while mitigating overfitting, although its MAE values were slightly higher than those of the embedding- based NN. The Decision Tree model showed reasonable accuracy ($R^2 \approx 0.80$), with low 7-day MAE but less stability at 28 days, reflecting its susceptibility to variance and limited generalization compared to ensemble approaches. Linear Regression, while computationally simple, yielded the weakest performance ($R^2 \approx 0.60$, MAPE > 10%), as it assumes linear relationships and cannot adequately capture the nonlinear hydration and interaction effects present in concrete mixtures. Transformer-based NN showed moderate results ($R^2 < 0.70$), suggesting that its advantages may not fully transfer to this structured tabular setting without further architectural and hyperparameter refinement.

Across all models, predictions at 28 days consistently outperformed 7-day predictions. This trend is consistent with the well- established behavior of concrete strength development, where early-age strength is influenced by variable hydration kinetics, supplementary cementitious material (SCM) reactivity, and curing conditions, leading to higher uncertainty. By 28 days, hydration and pozzolanic reactions are more advanced, reducing variability and allowing models to capture the strength more reliably. The substantial reduction in MAPE from 7-day to 28-day across all models supports this observation.

The scatter plots in Fig. 4 and Fig. 5 highlight the predictive capabilities of both traditional and machine learning models, with nonlinear methods consistently outperforming linear regression. Overall, the results highlight the potential of embedding-based neural networks for reliable prediction of field concrete strength, particularly in bridging the gap between early-age and standard acceptance strength assessments, while Random Forest provides a strong alternative where interpretability and computational efficiency are critical.

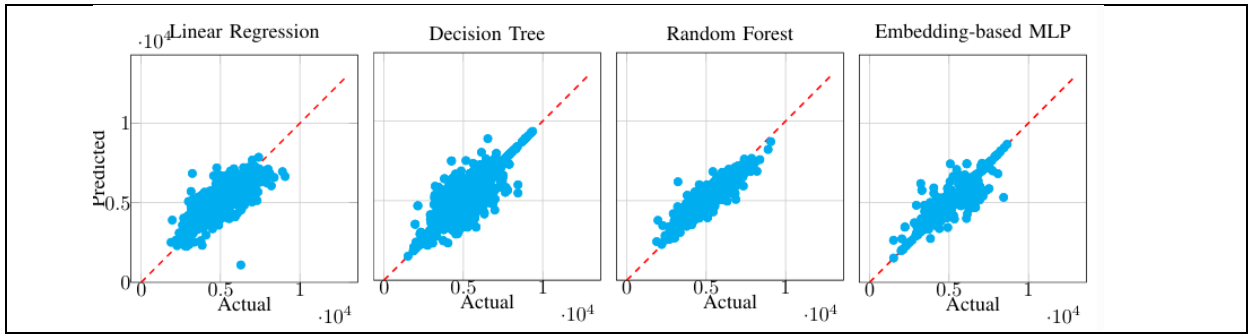

Fig. 4: Predicted vs Actual CCS for 7 days.

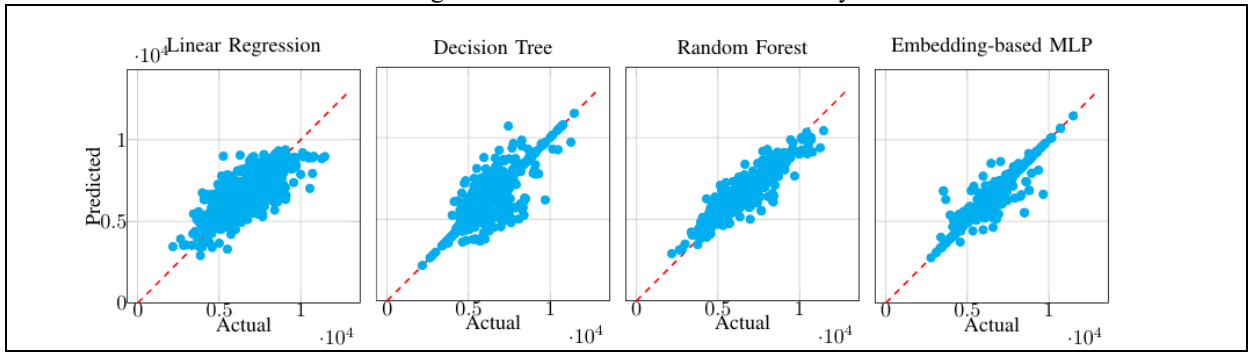

Fig. 5: Predicted vs Actual CCS for 28 days.

The feature importance analysis shown in Fig. 6 reveals distinct patterns across the evaluated models. Linear Regression places overwhelming weight on Material Code, reflecting its reliance on a strong linear association between categorical mix identifiers and compressive strength. Transformer-based models show a similar dependence on Material Code, though they also assign meaningful weight to the water-cement ratio and placement parameters, demonstrating their ability to capture sequence-like feature interactions. In contrast, Decision Tree and Random Forest models distribute importance more evenly across mix design variables, with unit cementitious material and actual water-cement (W/C) Ratio consistently ranked among the top contributors. This behavior underscores the advantage of ensemble methods in capturing nonlinear relationships and mitigating overfitting to a single dominant variable. Placement conditions and environmental factors, such as air content, slump, and temperature, exhibit moderate contributions, aligning with their known secondary influence on strength development. Notably, while tree-based and transformer models provide interpretable feature rankings, embedding-based neural networks do not offer inherent feature importance measures because their learning process projects categorical and numerical inputs into dense latent spaces where contributions are distributed across many parameters.

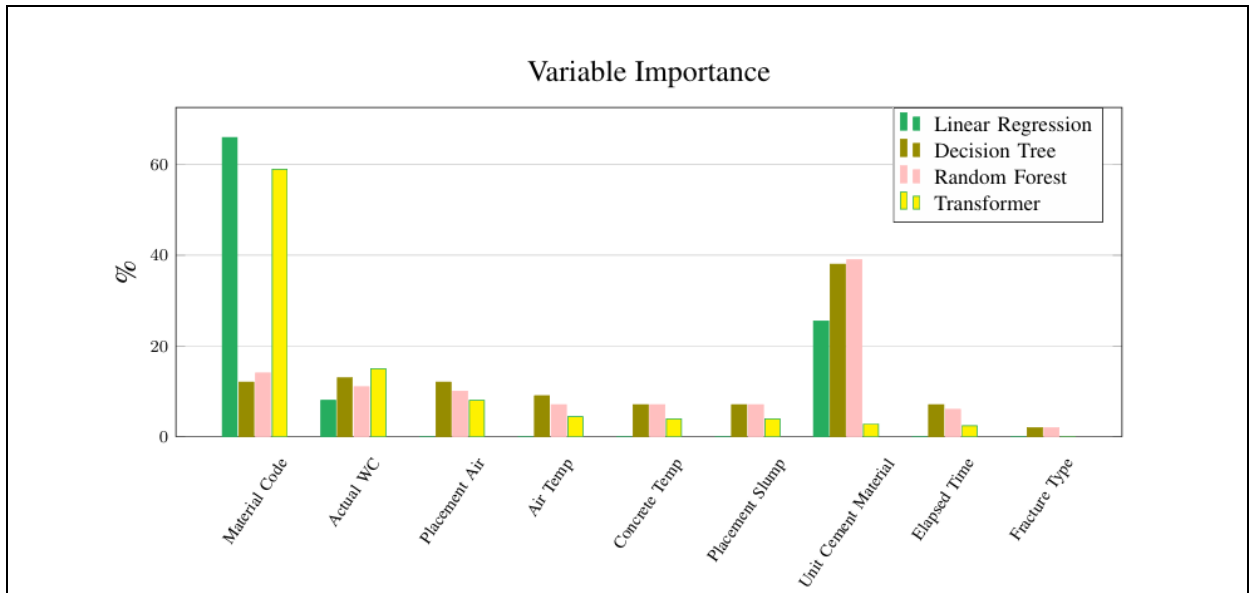

Fig. 6: Relative importance of the input variables.

## 6. Conclusion and Future Recommendation

The comparative evaluation demonstrated that the embedding-based artificial neural network yielded the most reliable performance for predicting concrete compressive strength using industry-scale field data. Its ability to capture complex interactions among categorical and numerical variables highlights its potential suitability for construction quality control applications. The 28-day compressive strength predictions achieved a mean absolute percentage error of approximately 2.5%, which is comparable to the magnitude of within-laboratory variability reported for standard compressive strength testing. According to ASTM C39/C39M, two properly conducted tests on standard-cured 100 × 200 mm (4 × 8 in.) concrete cylinders within the same laboratory may differ by up to 9% of their average strength [34]. This comparison provides context for the predictive accuracy achieved by the proposed model relative to routine laboratory testing uncertainty.

Future research should incorporate additional parameters such as curing conditions, ambient humidity, and admixture dosages to further improve predictive capability. Validation using datasets from diverse geographic regions and project types is also recommended to enhance generalizability. In addition, integrating data-driven strength prediction models within digital twin environments may support real-time analytics and proactive quality control for large-scale infrastructure projects. Beyond compressive strength, extending the framework to long-term performance indicators such as creep, shrinkage, and freeze–thaw resistance would further increase its relevance for performance-based construction specifications.